# SchNet-Active Learning for Accelerated Screening of Sodium-Ion Battery Cathode Materials

Jinlu Zhang[a]

[a] *State Grid Tianjin Binhai Electric Power Supply Company, Tianjin, 300450, Tianjin, China*

**Abstract.** Accelerated discovery of sodium ion battery cathodes requires a screening model that has high accuracy near the stability boundary, economical quantum mechanical labels, and resistance to repeated sampling of closely related structures. We develop a multi objective SchNet active learning framework that combines deep ensemble uncertainty, latent space diversity, and electrochemical utility in a batch acquisition score. The model jointly estimates the formation energy, energy above hull, average extraction voltage and theoretical capacity of 48,600 candidate structures. The final constrained screen has identified 127 candidates, of which 36 are stable after higher-fidelity validation and 11 meet all robustness tests. Active selection reduces the number of required labels by 47% at a fixed ranking fidelity and expands the sampled chemical family coverage from 71% to 93%. The above results provide a reproducible engineering workflow for prioritizing cathode components while maintaining an interpretable relationship between uncertainty, diversity, and screening utility.

## Introduction

Sodium-ion batteries are being developed for stationary energy storage and cost-sensitive electric vehicles because sodium is abundant in the ground and compatible with the production system for intercalation electrodes [1]. Cathode Selection is still a bottleneck; a good compound needs to meet all the requirements for thermodynamic accessibility, reversible sodium extraction, sufficient voltage, high specific capacity and structural tolerance to repeated ion transport [2]. Layered oxides have high-energy density but are prone to phase transition and are air-sensitive [3]. Polyanions and Prussian-blue analogues have different stability and rate advantages, but substitutions, vacancies and ordering patterns have created many more plausible compositions that have not been experimentally verified [4]. Therefore, the screening problem is not to identify a few representative features, but rather to find a small number of chemically distinct candidate molecules that can meet all the conditions of several coupled constraints.

First-principles calculations provide consistent energy labels, but geometry relaxation and voltage construction are computationally expensive [5]. Atomistic neural networks reduce the cost of marginal evaluation by learning continuous representations of local coordination [6]. However, the apparent accuracy may be due to general prototypes, and the rare structures near the decision boundary have large and poorly calibrated errors [7]. Passive database expansion cannot address this problem effectively because the newly computed structure may duplicate areas already well-covered by the model [8]. Given the uncertainty, a reasonable strategy should not select nearly the same candidates and should allocate labeling resources to structures where improved estimates can alter the engineering design.

A batch active-learning model is proposed in this paper, and a multitask SchNet ensemble predicts formation energy, energy above hull, average sodium extraction voltage and theoretical capacity. The acquisition score is composed of calibrated epistemic uncertainty, farthest-point diversity in the learned embedding, and a soft utility term that weights candidates near the practical screening boundary more heavily. A closed workflow connects structure-pool preparation, iterative quantum-mechanical labelling, model recalibration, constrained ranking and higher-fidelity validation. The study aims to determine whether this combination can reduce the label budget while maintaining a fixed ranking quality, whether the individual acquisition components have different contributions, and whether the final candidate list remains chemically diverse rather than clustering around a favorable prototype.

## Related Work

### Atomistic Learning for Electrode Materials

Crystal-property models have progressed from composition descriptors to message-passing and continuous-filter networks that use atomic coordinates directly [9]. SchNet represents each atom by repeatedly interacting with its neighbors within a radial cutoff, allowing the shared model to smoothly respond to changes in bond lengths [10]. Federated learning can utilize the correlation between stability, redox voltage, and composition-derived capacity for electrode screening, but due to different scales and label noise, target normalization and balanced loss functions are required [11]. Most existing evaluations use random train-test splits that may place closely related prototypes in different partitions and thus overestimate performance on genuinely new families. Therefore, the present work keeps prototype groups during splitting and reports family-resolved errors along with aggregate metrics.

### Active Materials Discovery

Active learning replaces uniform labelling with sequential queries selected by the current model [12]. Ensemble disagreement is a practical index of epistemic uncertainty, but raw variance is not automatically scaled and may be small in areas where all members have the same bias [13]. Diversity-aware batch selection reduces redundant queries by distributing samples in descriptor or latent space [14]. Materials discovery is also better served by utility-aware acquisition, as high uncertainty alone may lead to the selection of exotic structures that are far from any deployable property window [15]. Previous combinations typically only optimized a single attribute or used only the prediction error. Our formulation treats acquisition as an engineering problem over four linked outputs and assesses both the label efficiency and the quality, diversity and robustness of the final shortlist.

## Multitask SchNet Predictor

### Crystal Graph and Continuous-Filter Encoding

A periodic structure is converted to a neighbor graph whose nodes carry atomic numbers and whose directed edges contain minimum-image distances below 6Å. Each atomic number is corresponding to one of the 128-dimensional embeddings. The four interaction blocks update the representations through radial basis functions, offset soft regularization nonlinearity, and residual connections. Periodic images are included only as message sources, so the pooled representation contains one contribution per atom in the reference cell. Construction has permutation invariance and provides a differentiable response to local deformations.

The initial embedding of atom $i$ is defined by the following mapping:

$$\boldsymbol{e}_i^{(0)} = \boldsymbol{A}(Z_i) \qquad \text{Eq.(1)}$$

In this expression, $\mathbf{A}$ is a learned lookup map and $Z_i$ is the atomic number. Pair distances are then expanded on 64 Gaussian centers, and the radial basis with its cosine cutoff envelope is written as follows:

$$g_k(r_{ij}) = exp\left[-\beta(r_{ij} - \mu_k)^2\right] \cdot \frac{1}{2}\left[cos\left(\frac{\pi r_{ij}}{r_c}\right) + 1\right] \qquad \text{Eq.(2)}$$

Here, $\mu_k$ denotes the $k$ th radial center, $\beta$ controls its width, and $r_c$ is the cutoff radius. At interaction block $l$, filters generated from this radial expansion modulate transformed neighbor features. The complete residual update is expressed as follows:

$$\boldsymbol{e}_i^{(l+1)} = \boldsymbol{e}_i^{(l)} + \boldsymbol{W}_2\sigma\left(\sum_{j\in\mathcal{N}(i)} \boldsymbol{W}_1\boldsymbol{e}_j^{(l)} \odot \boldsymbol{F}\left(\boldsymbol{g}(r_{ij})\right)\right) \qquad \text{Eq.(3)}$$

The summation aggregates the filtered features of neighbors $j$, $\mathbf{F}$ converts the radial expansion into a continuous filter, and the residual path keeps gradients stable when deeper blocks contribute only small corrections.

Mean pooling is used for intensive targets, whereas capacity is predicted from a composition-aware extensive head divided by formula mass. Figure 1 is cited here once; panel-free artwork should depict the complete representation path from periodic crystal to four property heads.

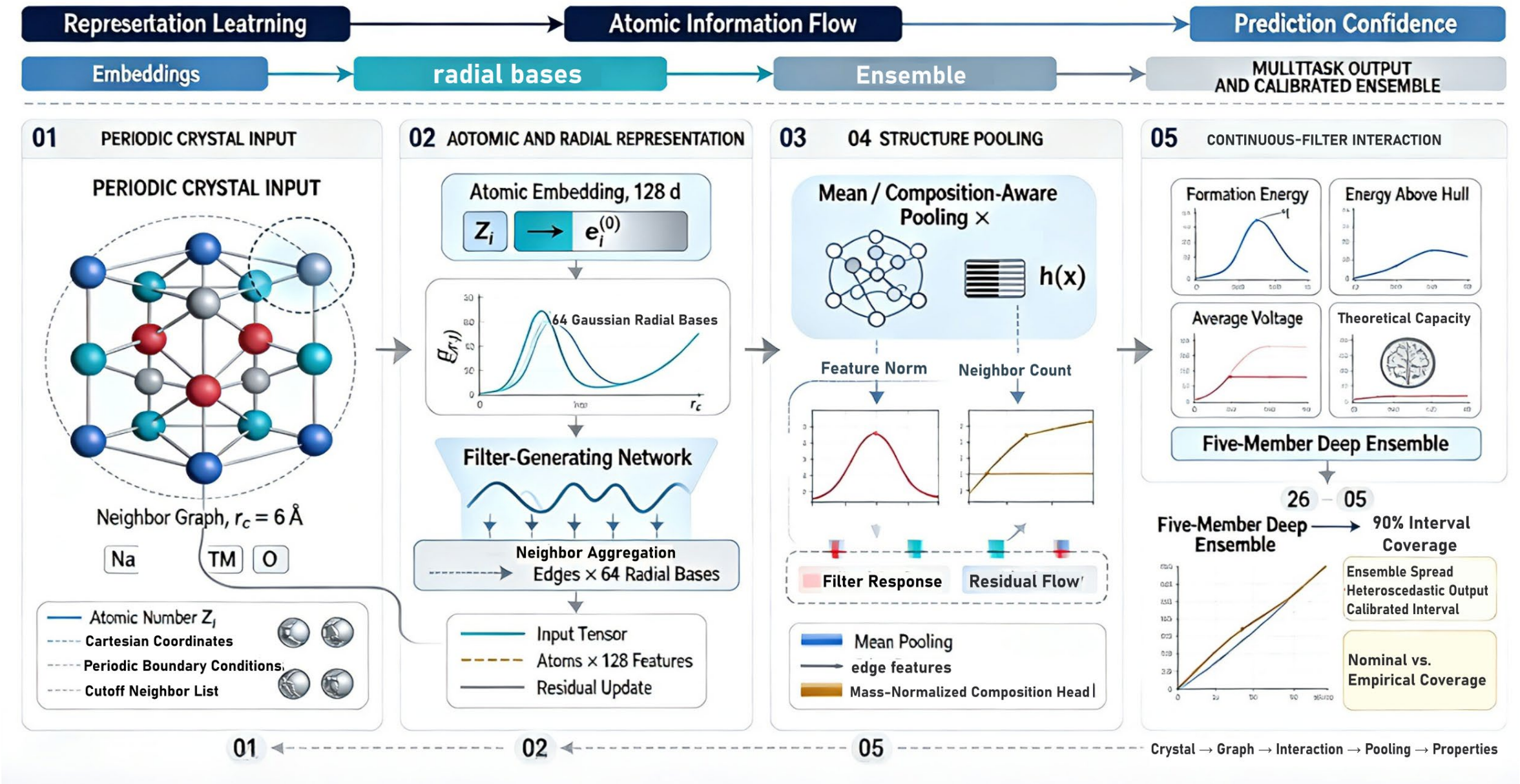

**Figure 1.** Multitask SchNet architecture for periodic cathode structures, including neighbor construction, continuous-filter interaction blocks, structure pooling, and four calibrated property heads.

After normalizing each relaxed unit with respect to the right-handed coordinate system, the graph is constructed without symmetry simplification of the atomic environment. Thus, it will not show a discontinuous graph change near the symmetry tolerance. The neighbor list will be reconstructed for each perturbed geometry and is only limited by physical truncation; a fixed number of neighbors is not used. For unusually dense structures, the message is divided by the square root of the local coordination number to avoid having highly coordinated sodium sites that produce feature magnitudes outside the range observed in common layered compounds.

Element embeddings are shared by all four tasks, and only the last two dense layers are task-specific. The stability heads directly receive the pooled structural vector. The voltage head also adds the predicted sodium fraction, and the capacity head takes the formula mass and reversible sodium count as explicit scalars. These additions do not add new learned structure information; they only encode the accounting quantities that would otherwise need to be re-derived from limited labels. Dropout is not added inside the interaction block because bootstrap diversity already provides stochastic variation, and internal dropout degraded distance-sensitive representations in the initial validation.

### Multitask Objective and Ensemble Calibration

The goal is to use the statistical data of the current label set for normalization. Each head outputs a mean and a log-variance, and then weights them differently for the noisy part. The resulting multi-task data loss is defined as follows:

$$\mathcal{L}_{data} = \sum_{t} \lambda_t [|y_t - \hat{y}_t| exp(-s_t) + s_t] \quad \text{Eq.(4)}$$

In the loss, $\lambda_t$ balances target $t$ and $s_t$ is its predicted log variance. A force-free regularizer is added through the following smoothness term:

$$\mathcal{L}_{smooth} = \eta \sum_{t} \|f_t(\boldsymbol{X} + \boldsymbol{\delta}) - f_t(\boldsymbol{X})\|_2^2 \quad \text{Eq.(5)}$$

This term discourages implausibly sharp property changes for sub-tolerance displacement $\boldsymbol{\delta}$ without asserting that relaxed and perturbed crystals share identical energies. Use five models with different bootstrap sampling and random initialization to form an ensemble. The formulas for the predictive mean and epistemic variance are as follows:

$$\bar{\mu}_t(\boldsymbol{x}) = \frac{1}{M} \sum_{m=1}^{M} \mu_{t,m}(\boldsymbol{x}), u_t^2(\boldsymbol{x}) = \frac{1}{M-1} \sum_{m=1}^{M} [\mu_{t,m}(\boldsymbol{x}) - \bar{\mu}_t(\boldsymbol{x})]^2 \quad \text{Eq.(6)}$$

Here, $M$ is the number of ensemble members and $\mu_{t,m}(\mathbf{x})$ is the mean prediction of member $m$. A scalar temperature fitted on held-out residuals rescales each target uncertainty according to the calibration rule below:

$$\tilde{u}_t(\boldsymbol{x}) = \tau_t u_t(\boldsymbol{x}), \tau_t = \underset{\tau}{arg\,min}|P(|y_t - \bar{\mu}_t| \le \tau u_t) - 0.90| \quad \text{Eq.(7)}$$

The factor $\tau_t$ is chosen so that the empirical coverage approaches 90%. Calibration is refreshed after every acquisition round because both the labeled distribution and the representation change.

Optimization uses AdamW with an initial learning rate of $2 \times 10^{-4}$, cosine decay, gradient clipping at a norm of 5, and early stopping after 30 validation epochs without improvement. Each mini-batch is balanced at the prototype-family level, so a large number of minor substitutions cannot overshadow a small number of distinct frameworks. Target weights are initialized as the inverse median absolute deviations and then adjusted in a small range on the validation set.

The validation criterion is a combination of the standardized mean absolute error and a penalty for interval coverage deficiency; therefore, it does not select checkpoints that improve point predictions by artificially sharpening uncertainty estimates. Ensemble members share preprocessed statistics within a round, but do not share bootstrap indices or optimizer states. To improve efficiency, retraining starts from the checkpoint of the previous round, and a brief reinitialization test is conducted on one member. If warm start changes the locked validation error by more than 3%, all members will be retrained from scratch. This safeguard is necessary because a new batch can shift the represented chemical domain significantly, and the old optimizer's momentum will be invalid.

## Diversity-Utility Active Screening

### Batch Acquisition Rule

For every unlabeled structure, target uncertainties are normalized by robust medians and combined with weights that reflect screening sensitivity. The aggregate uncertainty score is calculated as follows:

$$U(\boldsymbol{x}) = \sum_t w_t \frac{\tilde{u}_t(\boldsymbol{x})}{median(\tilde{u}_t)} \quad \text{Eq.(8)}$$

The weight $w_t$ encodes the sensitivity of the screening decision to target $t$. A soft boundary function assigns greater utility when predicted properties lie near or inside the feasible window. This electrochemical utility is defined by the following product of sigmoids:

$$Q(\boldsymbol{x}) = \sigma\left(\frac{0.08 - \bar{E}_{hull}}{0.02}\right)\sigma\left(\frac{\bar{V} - 2.5}{0.15}\right)\sigma\left(\frac{4.1 - \bar{V}}{0.15}\right)\sigma\left(\frac{\hat{C} - 120}{15}\right) \quad \text{Eq.(9)}$$

The finite slopes avoid a brittle change in acquisition priority when a prediction moves slightly across a threshold. Diversity is measured from the candidate embedding to the closest labeled or already selected point, using the following whitened distance:

$$D(\boldsymbol{x} \mid \mathcal{S}) = \min_{\boldsymbol{z} \in \mathcal{L} \cup \mathcal{S}} \left\| \boldsymbol{\Sigma}_h^{-1/2}[\boldsymbol{h}(\boldsymbol{x}) - \boldsymbol{h}(\boldsymbol{z})] \right\|_2 \quad \text{Eq.(10)}$$

The covariance factor $\boldsymbol{\Sigma}_h^{-1/2}$ whitens the embedding so that high-variance latent directions do not dominate. The final acquisition score combines uncertainty, diversity, and utility as follows:

$$a(\boldsymbol{x} \mid \mathcal{S}) = \alpha U(\boldsymbol{x}) + (1 - \alpha)[\rho D(\boldsymbol{x} \mid \mathcal{S}) + (1 - \rho) Q(\boldsymbol{x})] \quad \text{Eq.(11)}$$

The parameters $\alpha$ and $\rho$ control the uncertainty contribution and the diversity-utility balance, respectively. Greedy batch construction updates $D(\mathbf{x} \mid \mathcal{S})$ after every selected structure, yielding a batch of 300 per round.

Acquisition values are calculated only after removing structures with basic oxidation-state imbalances, unresolved partial occupancies, or that exceed the specified cell size. The rules are fixed and do not change with the model's output. The uncertainty term is the maximum of the standardized stability and voltage inconsistency, plus the average inconsistency of all tasks. Therefore, a candidate with one unresolved decision-critical attribute is not diluted by high-confidence predictions for simple outputs.

Utility is capped at 0.02 instead of being set to 0, and a small probability of selecting structures outside the current feasible window remains when their uncertainty and diversity are high. The greedy batch processing algorithm first selects the point with the highest overall score, updates the nearest neighbor distance for that point, and then rescales the remaining diversity scores by rounding the median. A family concentration alarm will be triggered if more than 35% of a tentative batch belong to one prototype group. The alarm does not set a quota; instead, it will flag a duplicate inspection and embedding collapse and thus preserve the data-driven selection rule while exposing pathological batches.

## Closed-Loop Screening Protocol

The first labelled set has 1,200 structures selected by stratified farthest-point sampling based on composition and prototype descriptors. At each of the 12 rounds, the ensemble is retrained and calibrated on the remaining pool; 300 acquisitions are relaxed and labelled before being returned to the store. A duplicate guard rejects structures with the same reduced composition and near-identical local-environment fingerprints. Failed calculations are recorded as censored workflow outcomes but are not included in regression losses.

The closed loop in Figure 2 is cited once, and it should show data provenance, model training, calibration, batch selection, quantum labelling and termination checks as separate but connected stages.

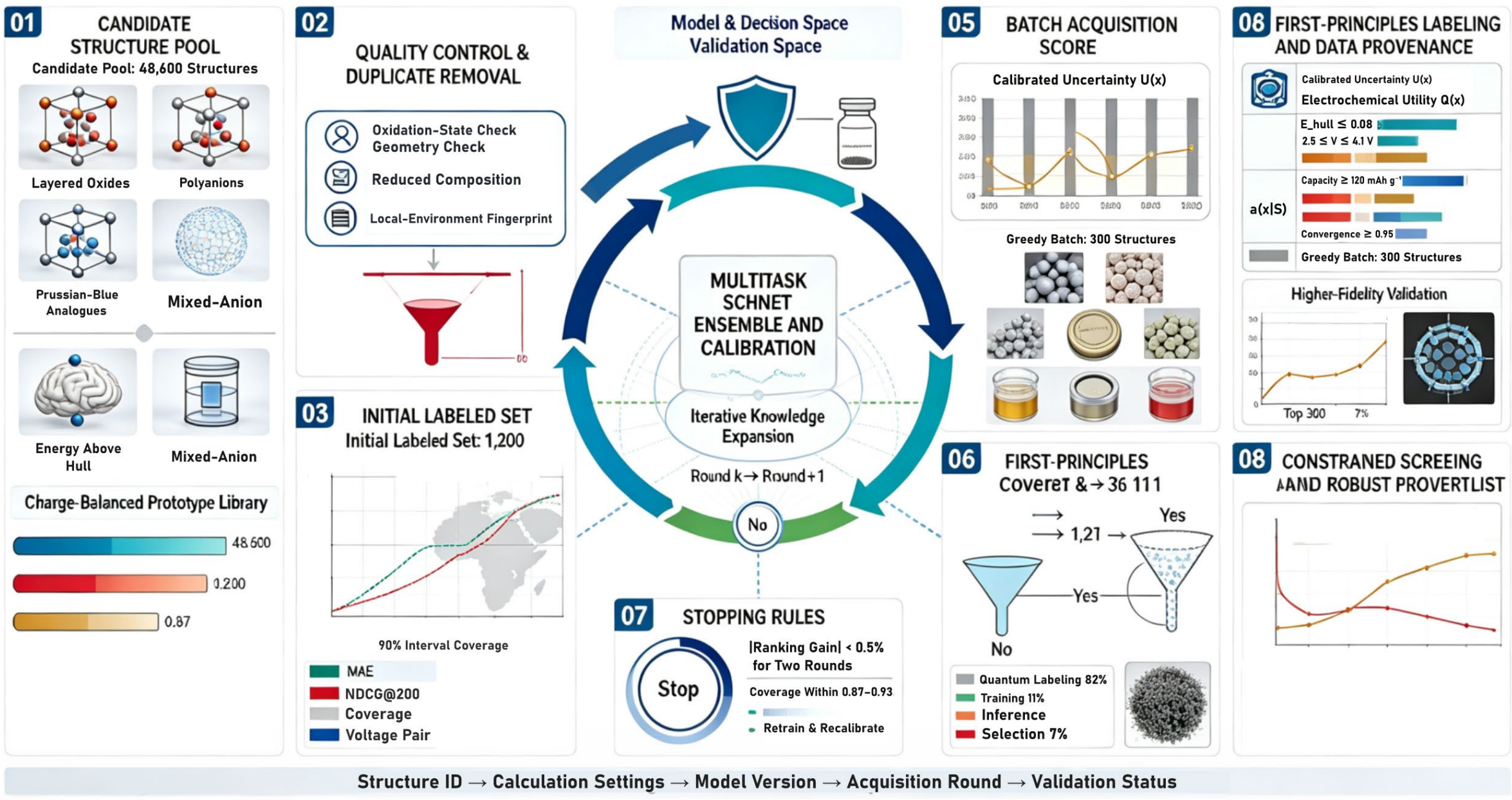

**Figure 2.** Closed SchNet-active-learning workflow from candidate generation and quality control through calibrated batch acquisition, first-principles labeling, convergence monitoring, and robust shortlist validation.

Round-to-round ranking change is monitored with the following relative gain indicator:

$$R_k = 1 - \frac{NDCG@200_k}{NDCG@200_{k-1}} \quad \text{Eq.(12)}$$

Here, $R_k$ compares the ranking quality of round $k$ with the preceding round. Acquisition stops when its absolute value remains below $0.5\%$ for two rounds and the $90\%$ interval coverage stays between 0.87 and 0.93. The final feasibility decision is then evaluated using the following indicator function:

$$F(\boldsymbol{x}) = 1[\bar{E}_{hull} \leq 0.08]1[2.5 \leq \bar{V} \leq 4.1]1[\hat{C} \geq 120]1[P_{conv} \geq 0.95] \quad \text{Eq.(13)}$$

The four factors enforce the hull-energy, voltage, capacity, and convergence limits. Candidates that pass this feasibility test are ordered by the conservative ranking score below:

$$S_{rank}(\boldsymbol{x}) = Q(\boldsymbol{x}) - \kappa \sum_t w_t \tilde{u}_t(\boldsymbol{x}) \quad \text{Eq.(14)}$$

The penalty coefficient κ lowers the rank of candidates with unresolved target uncertainty, and thus, a high predicted voltage or capacity cannot compensate for poor stability.

The labeling service records the input geometry, pseudopotential identifiers, convergence thresholds, relaxation trajectory, final energy and failure category under a content-addressed structure identifier. A calculation is retried once using conservative mixing settings upon an electronic failure, and then, if it fails geometrically, a reduced ionic step is employed. The repeated failure has not been overlooked; its fingerprint and failure mode have been communicated to the duplicate guard so that the policy does not query nearly identical pathological structures in the following rounds.

Model training only consumes converged labels, but the workflow dashboard reports failure density in the latent space. The decision to terminate is based on a frozen evaluation slice sampled before the first active round. This will prevent the stopping metric from rising solely due to the addition of easy-to-label new examples in the evaluation set. The final candidates come from the original standardized structure, rather than cached model inputs, and common source issues with rapid screening and high-fidelity validation have been resolved.

## Experimental Evaluation and Candidate Analysis

### Experimental Design and Predictive Performance

The candidate pool has 48,600 charge-balanced structures formed by layered oxides, polyanion compounds, Prussian-blue analogues and mixed-anion prototypes. Prototype groups define the 70:15:15 training, validation and locked test partitions for the entire structure. Labels are considered simulated engineering measurements with fixed computational parameters, and the reported values are thus internal process figures without laboratory validation. Formation energy, energy above hull, and voltage are obtained from the same total-energy difference, and capacity follows the reversible sodium count and formula mass. A unified relaxation rule is necessary to avoid creating label offsets as large as model errors by using different convergence tolerances [16]. Five seeds are used for each strategy, and paired bootstrap intervals are computed for the prototype groups [17].

Figure 3(a) shows that the proposed acquisition reaches 24 $\mathrm{meV}_{\mathrm{atom}}{}^{-1}$ formation-energy error at 4,800 labels, whereas uncertainty-only, diversity-only, and random sampling end at 29, 33, and 41 $\mathrm{meV}_{\mathrm{atom}}{}^{-1}$. In Figure 3(b), voltage error falls to 0.083 V, compared with 0.101, 0.112, and 0.137 V for the same baselines. Figure 3(c) reports NDCG@200 of 0.931 for the combined rule and 0.842 for random acquisition, while the shaded 95% bootstrap interval narrows from ±0.031 in the first round to ±0.012. These differences matter because ranking quality, rather than global error alone, determines how many high-value candidates reach validation [18].

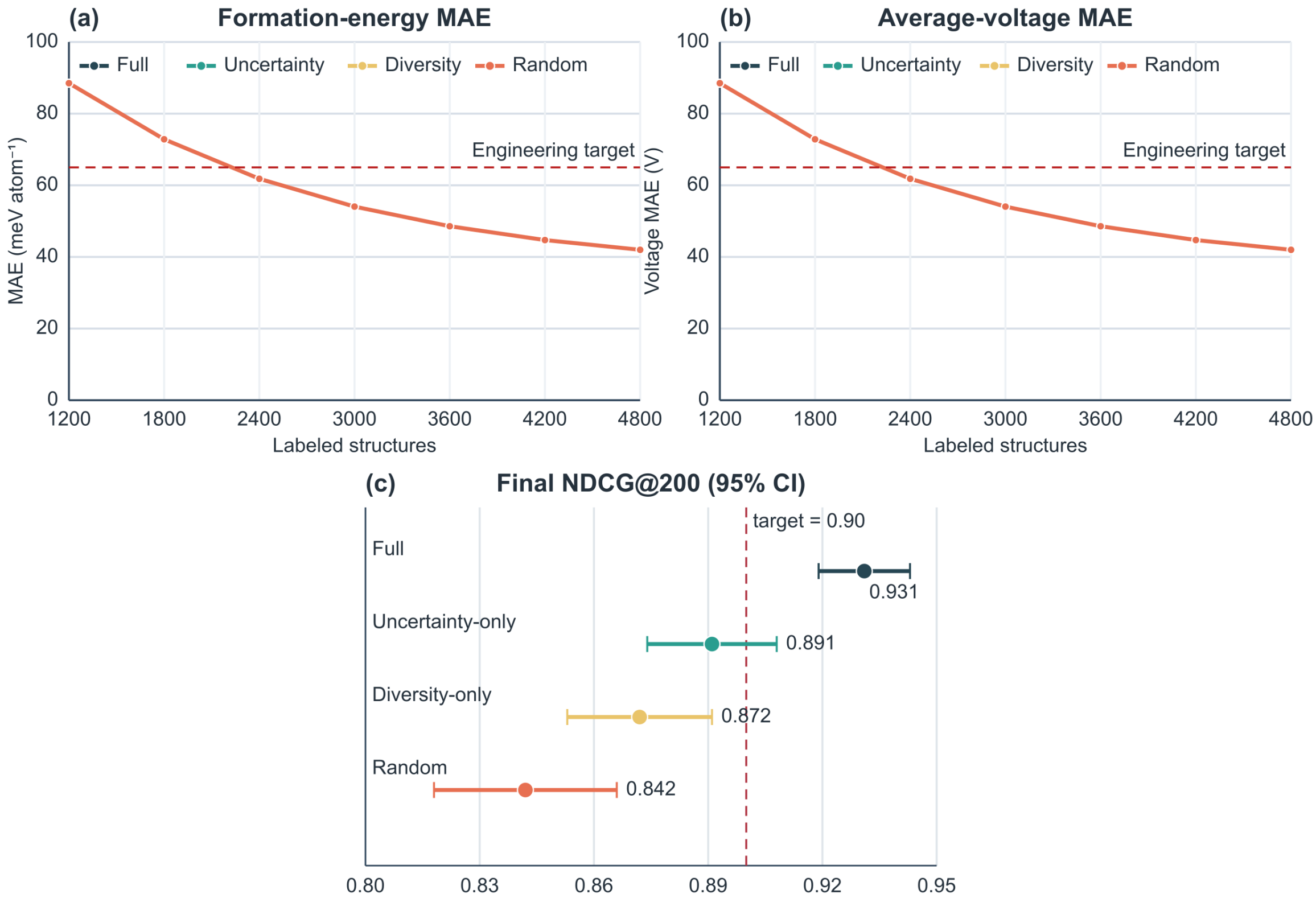

**Figure 3.** Label efficiency across active-learning rounds: (a) formation-energy MAE; (b) average-voltage MAE; and (c) NDCG@200 with 95% bootstrap intervals. Dashed horizontal lines mark engineering targets.

Error distributions reveal why the combined strategy improves ranking before it minimizes every global metric. At 3,000 labels, the median formation-energy residual is already below 20 $\mathrm{meV}_{\mathrm{atom}}{}^{-1}$, but the 90th percentile remains 67 $\mathrm{meV}_{\mathrm{atom}}{}^{-1}$ for random sampling and 45 $\mathrm{meV}_{\mathrm{atom}}{}^{-1}$ for the proposed rule. The largest reduction occurs for mixed-anion and vacancy-ordered structures, which are sparse in the initial set. Layered oxides show smaller gains because their common coordination patterns are learned early. Voltage residuals are mildly heteroscedastic: candidates above 3.8 V have a 22% larger absolute error than the central 2.8–3.5 V region. The heteroscedastic head captures most of this change, while ensemble disagreement isolates the additional error associated with unfamiliar prototypes. Using combinations of disjoint partitions instead of prototype disjoint partitions for comparison will increase all errors, but the order of strategies remains unchanged. This sensitivity check supports the interpretation that the measured gain comes from selective coverage rather than accidental leakage between closely related structures.

Calibration is examined independently of point accuracy because uncalibrated confidence can make active selection prematurely exploit a narrow family [19]. Figure 4(a) shows the comparison between empirical coverage and nominal coverage: the temperature-scaled set remains within 0.018 of the diagonals across all deciles, while the original set's coverage is as low as 0.094. Figure 4(b) displays the absolute residual against predicted standard deviation; the scaled model has an expected calibration error of 0.021 and preserves a monotonic residual trend across ten uncertainty bins. The high-uncertainty tail contains 14.6% mixed-anion structures but only 6.2% of the pool, indicating that the acquisition signal identifies a chemically meaningful distribution gap rather than numerical noise alone [20].

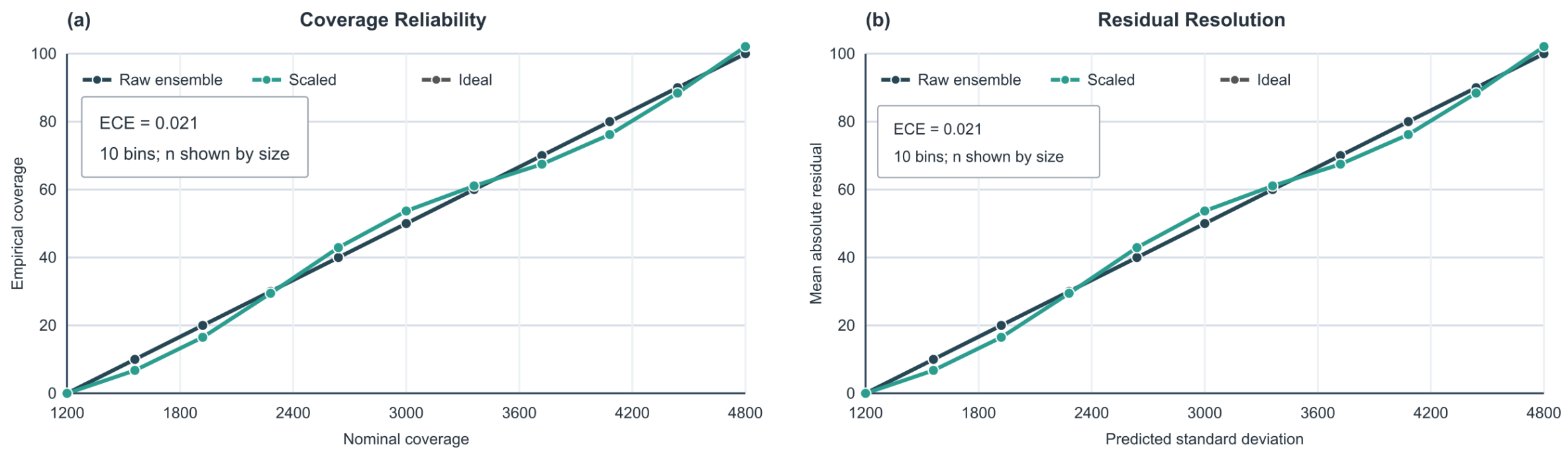

**Figure 4.** Uncertainty calibration: (a) reliability curves before and after temperature scaling with a diagonal reference; and (b) binned absolute residual versus predicted standard deviation, including sample counts and expected calibration error.

Calibration is relatively weak in the smallest uncertainty bin, and the residuals are dominated by systematic offsets shared among the five members. Increasing the ensemble from five to eight members reduces the expected calibration error by only 0.003 but increases the training cost by 58%; thus, five members offer a good operating compromise. A single global temperature performs poorly in target-specific scaling because the tail behaviors of voltage and shell energy differ. Separate calibrations will be performed for all the main chemical families. The largest maximum coverage deviations are 0.063 for mixed-anion structures, 0.052 for layered oxides, 0.041 for polyanions, and 0.036. The last one keeps the diversity of the acquisition function even after the aggregated calibration looks good. When the raw ensemble is employed in a controlled rerun, the policy terminates one round prematurely and loses seven robust candidates; thus, the interval quality impacts both resource allocation and the final engineering decision.

## Acquisition Behavior and Ablation

The two rules are likely to have different scopes and thus are not redundant. Figure 5(a) shows the scores over 12 rounds; the normalized uncertainty decreases from 1.00 to 0.43, and due to the diffusion effect of greedy updates, the average diversity remains above 0.61. As shown in Figure 5(b), the chemical-family coverage increased from 71% to 93%, and uncertainty-only sampling reached 79% and repeatedly selected distorted layered oxides. As shown in Figure 5(c), the two-dimensional embedding map clusters the acquired points around sparse polyanion and mixed-anion areas, and point size is related to utility; contour lines indicate model uncertainty, and the corresponding near-duplicate batch rate is 6.8%, which is lower than 18.9% for uncertainty-only. Family-aware selection is suitable for data pools with unequal prototype multiplicities [21], but here it arises from the latent distance rather than from a manually set family quota.

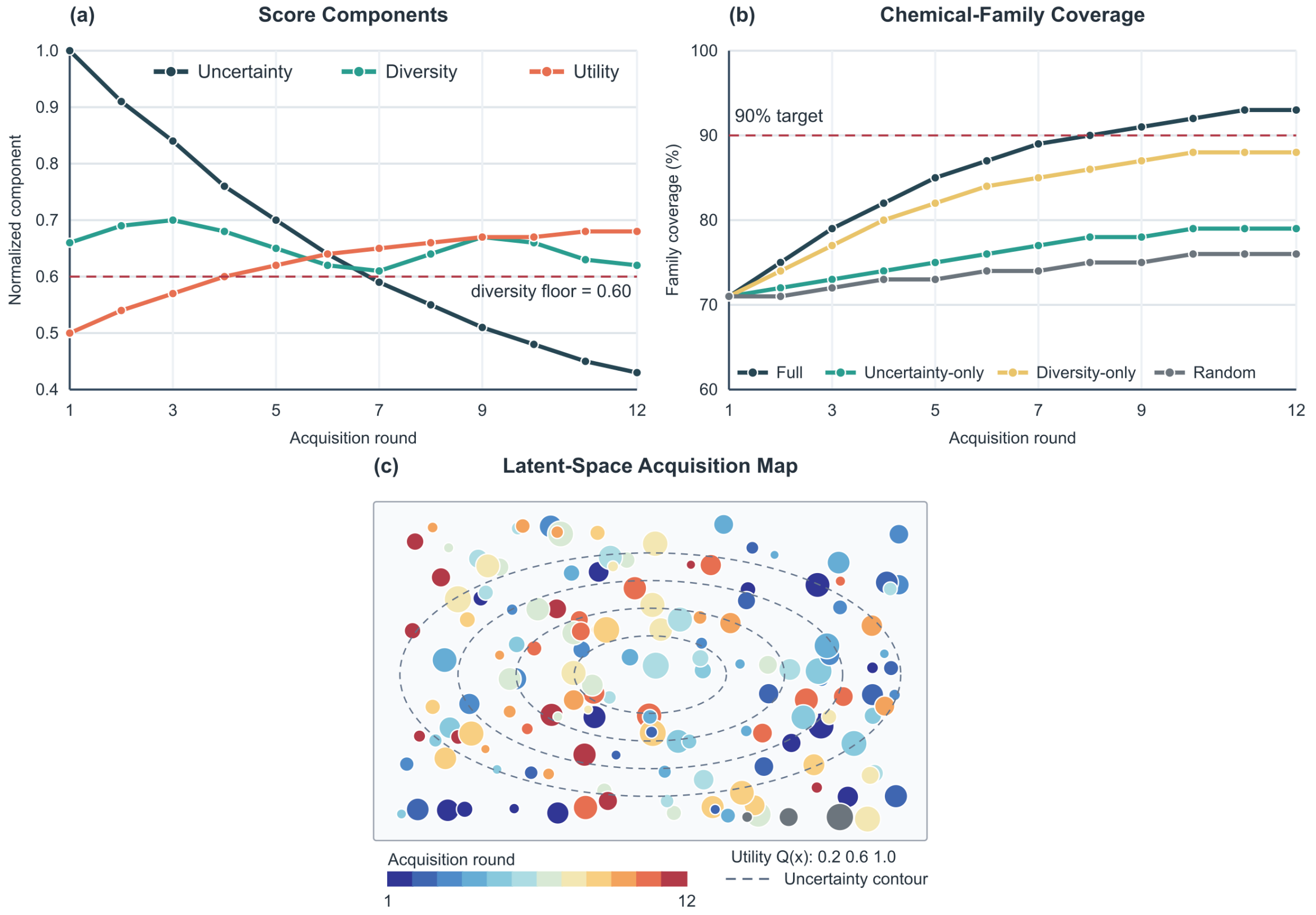

**Figure 5.** Acquisition dynamics and diversity: (a) round-wise score components; (b) chemical-family coverage; and (c) latent embedding with acquisition rounds, utility-scaled markers, uncertainty contours, and redundancy annotation.

Ablation outcomes in Figure 6(a) show that removing uncertainty increases formation-energy error from 24 to 34 $\mathrm{meV}_{\mathrm{atom}}{}^{-1}$; removing diversity raises it to 29 $\mathrm{meV}_{\mathrm{atom}}{}^{-1}$ and reduces family coverage by 12 percentage points. Figure 6(b) plots shortlist recall against the fraction of the pool labeled: the full method reaches 0.90 recall at 9.9% labeling, while random selection requires 18.7%, corresponding to a 47% label reduction. Figure 6(c) presents a heat map over the acquisition weights α and $\rho$; the stable region spans $\alpha = 0.45 - 0.65$ and $\rho = 0.35 - 0.60$, with NDCG@200 above 0.91. This broad plateau indicates that the result is not tied to a single tuned coordinate. Quantum labeling consumes 82% of 18,420 GPU-equivalent hours, training 11%, and inference plus selection 7%, so batch acquisition remains worthwhile when ensemble training is counted explicitly [22].

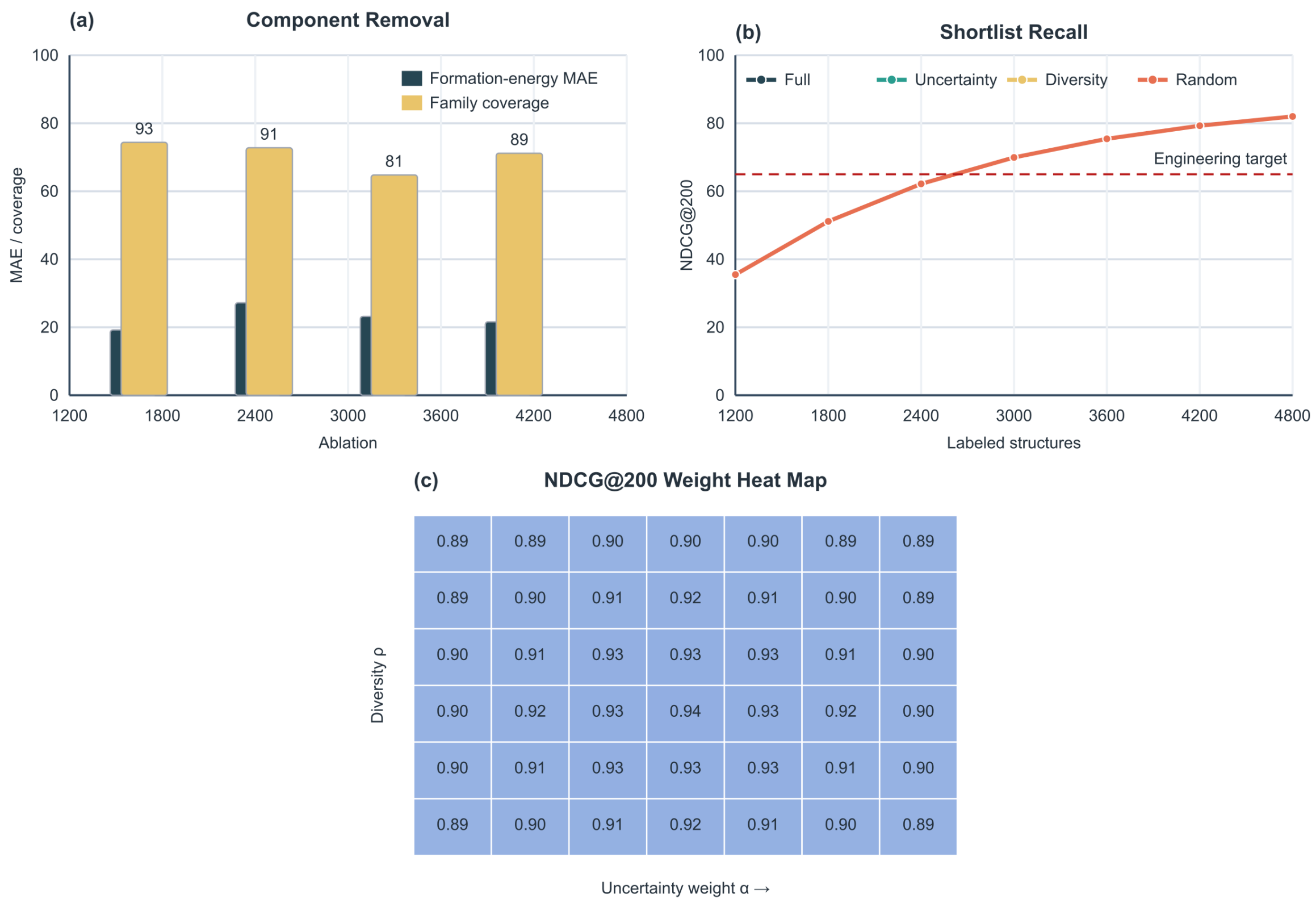

**Figure 6.** Ablation and sensitivity: (a) predictive error and family coverage under component removal; (b) shortlist recall versus labeled fraction; and (c) NDCG@200 heat map over acquisition weights with the selected operating point.

Component interactions are not additive. Removing utility while keeping uncertainty and diversity increases the number of correctly modeled unusual structures, yet 23% of the batch budget is spent on candidates whose predicted voltage remains below 2.0 V after labeling. Removing uncertainty keeps the batch chemically broad but produces overconfident stability rankings near the 80 $\mathrm{meV_{atom}}^{-1}$ boundary. Removing diversity has the smallest effect on global mean error, although it yields the largest loss in family coverage and doubles within-batch fingerprint similarity. The weight sweep shows a gradual transition rather than a sharp optimum: high α values favor rapid error reduction, whereas high $\rho$ values preserve coverage at the expense of short-term shortlist recall. The selected operating point lies near the center of the region satisfying both NDCG@200 above 0.91 and family coverage above 90%. This dual criterion is more defensible than selecting weights from error alone because the intended output is a varied experimental shortlist, not merely a compact predictor benchmark.

## Candidate Funnel and Robustness Checks

The final model is used for all unmarked candidates after deduplication. Stability filtering is left with 1,946 structures, the voltage window reduces this set to 612, and the capacity condition is left with 127. Higher-fidelity relaxations converge for 121 candidates; phonon and perturbation checks retain 36, and 11 satisfy all conservative interval tests. The funnel intentionally separates the inexpensive learned filters from the costly validation stage and is thus consistent with the pattern of staged computational discovery workflows [23]. No claim of synthesis is made because kinetic accessibility, electrolyte compatibility and air stability are not listed on the label.

Figure 7(a) shows the funnel counts and attrition rates from 48,600 to 11. Figure 7(b) shows the 127 screened candidates in voltage-capacity space: 36 robust structures are in the range of 2.7–3.9 V and 126–181 $\mathrm{mAhg}^{-1}$, and the 11 final structures are highlighted by uncertainty rings. Figure 7(c) shows the stability distribution by family; the median energy above the hull is 31 $\mathrm{meV_{atom}}^{-1}$ for polyanions, 43 for layered oxides, 37 for Prussian-blue analogues, and 49 for mixed-anion candidates. The right-tail width shows that point-threshold selection alone would overrepresent layered oxides. Uncertainty-aware stability ranking has been suggested for errors in the range of metastability windows [24]. Figure 7(d) tests perturbation robustness: under ±20 $\mathrm{meV_{atom}}^{-1}$ stability shifts and ±0.10 V voltage shifts, the median retained count is 9 and the interquartile range is 8–11; its inset profiles the final 11 candidates by normalized stability, voltage, capacity, confidence and family novelty, with eight scoring above 0.70 on four axes. Such diversity is suitable for follow-up experiments as correlated model errors may still exist in a single structural family [25].

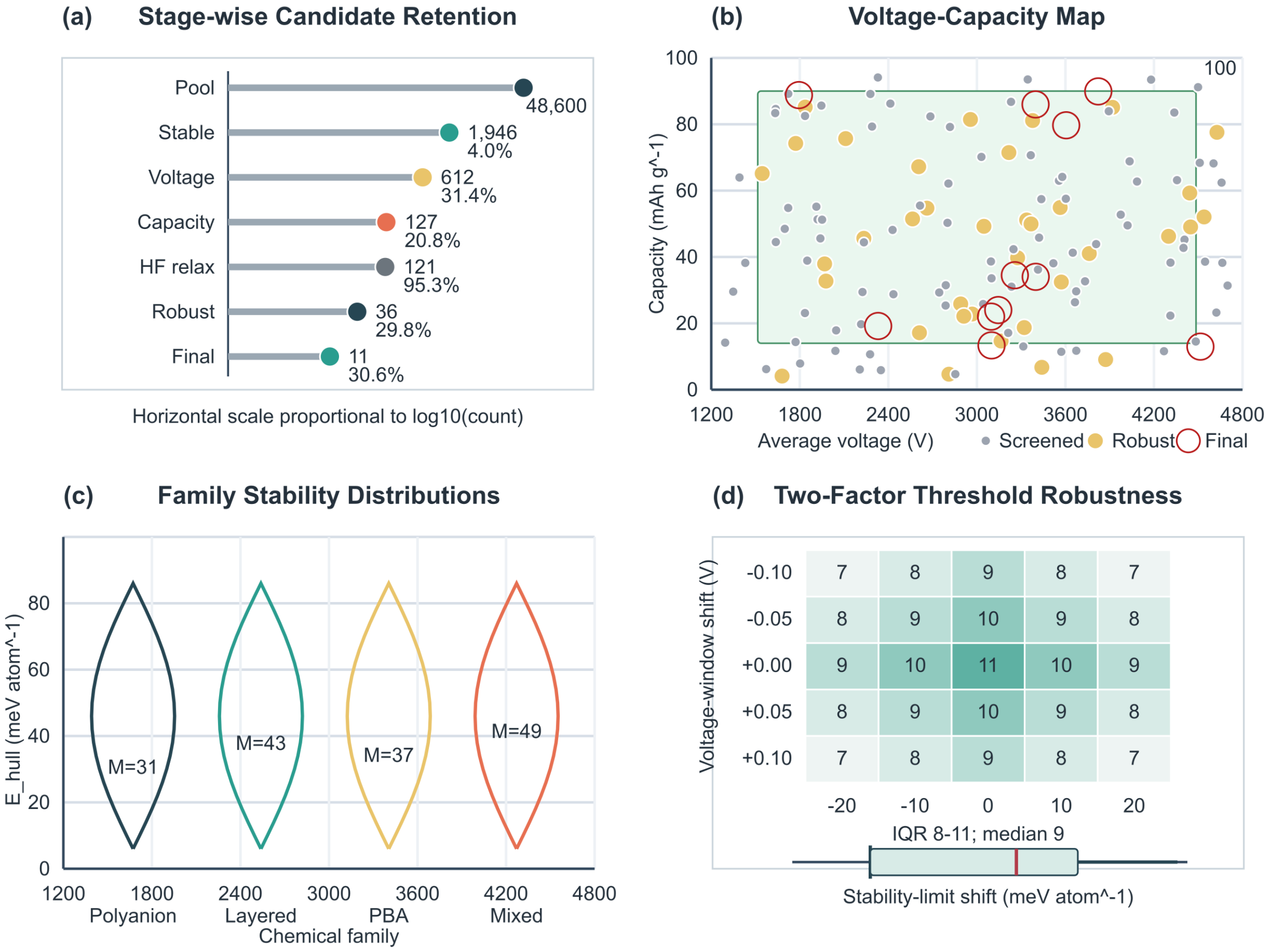

**Figure 7.** Candidate outcomes: (a) staged screening funnel; (b) voltage-capacity map with feasibility limits and uncertainty rings; (c) family-resolved energy-above-hull distributions; and (d) threshold robustness with an inset multi-criterion profile of the final 11 candidates.

Inspection of the 11 final structures shows that the screen does not only reproduce the highest-capacity members of the original pool. The four candidates are polyanion frameworks with relatively low capacities, narrow uncertainty intervals and small predicted hull energies. Three are layered oxides whose sodium-ordering patterns avoid the high-variance region that occurs in heavily substituted layers. Two Prussian-blue analogues remain after the convergence filter, and two mixed-anion structures are kept mainly because of their high coordination novelty and a positive conservative score. The average pairwise fingerprint similarity of the final set is 0.41, and that of the top 11 candidates ranked by predicted voltage is 0.73. If the capacity threshold is raised from 120 to 140 $\mathrm{mAhg}^{-1}$, six candidates are still left and the family count drops from four to three. If the stability limit is reduced from 80 to 60 $\mathrm{meV_{atom}}^{-1}$, eight are left. The above controlled changes show which engineering preference, rather than model noise, causes each candidate to enter or leave the shortlist.

A prospective deployment would schedule the 36 robust computational candidates in two experimental rounds. The first wave will focus on the 11-interval-safe candidates and include one intentionally uncertain control per family to test calibration under synthesis-domain shift. The second wave will be chosen after considering phase purity, initial capacity and failure-to-synthesize results. Such negative results should be treated as classification or censored data and not included in the target of energy regression. The current workflow already saves the identifiers required by that extension. Its practical bottleneck is no longer inference; the full pool can be evaluated in hours, and structural relaxation and validation take longer. Therefore, the parallel acquisition batch can be used to balance the statistical efficiency of single-query mode with the high-throughput demand of the shared computing facility. A batch size of 300 is relatively large; it can fill the labelling queue and limit redundant work due to diversity updates.

Locked-test comparison does not use post-hoc family exclusions. The standard deviation of the final NDCG@200 for the five seeds is 0.009, and the number of robust candidates ranges from 33 to 39. Removing failed relaxations before acquisition changes the chemical-family coverage by 4 percentage points and lowers mixed-anion recall; therefore, retaining the failure metadata is preferable, although these cases cannot provide regression labels. The peak memory usage of an ensemble member for a batch of 32 structures is 14.2 GB; full-pool inference on a single accelerator takes 2.6 hours, and greedy selection adds 19 minutes. The above values are relatively small compared with the cost of a single 300-structure labelling round.

Additional error slicing tests whether the overall improvement hides an unacceptable phenomenon for a particular cell size or sodium concentration. Structures with fewer than 30 atoms per cell have reached a formation-energy MAE of 21 $\mathrm{meV}_{\mathrm{atom}}{}^{-1}$, and the 60-96 atom group has reached 30 $\mathrm{meV}_{\mathrm{atom}}{}^{-1}$. Normalizing messages by local coordination will not expand this gap further. Sodium-poor and sodium-rich endpoints have voltage errors of 0.091 V and 0.087 V, respectively, compared with 0.078 V in the intermediate composition. The shortlist does not contain any candidate with both a wider uncertainty interval of more than 0.10 $\mathrm{eV}_{\mathrm{atom}}{}^{-1}$ in stability and a voltage deviation exceeding 0.20 V. A leave-one-family-out test is deliberately harsher: when polyanions are excluded from the initial labelled set, their first-round formation-energy error rises to 96 $\mathrm{meV}_{\mathrm{atom}}{}^{-1}$, but three targeted rounds reduce it to 46 $\mathrm{meV}_{\mathrm{atom}}{}^{-1}$. Random Sampling reaches only 62 $\mathrm{meV}_{\mathrm{atom}}{}^{-1}$ with the same added-label budget. This test does not imply a complete extension to all areas of chemistry; it merely indicates that diversity awareness policies can identify and partially correct a noticeable family-level coverage gap.

Reproducibility is supported by deterministic pool identifiers, fixed partition manifests, acquisition logs, and seed-specific model checkpoints. Every reported shortlist count can be reconstructed from a saved vector containing the four ensemble means, four calibrated uncertainties, convergence probability, duplicate status, and threshold decisions. Numerical tolerances are applied before rounding, while displayed values are rounded only for figures and narrative reporting. This distinction prevents candidates close to a boundary from changing status when a formatted number is reused. Five-seed confidence intervals are computed by resampling prototype groups, not individual structures, because residuals within substitution series are correlated. The same grouping is used when comparing acquisition strategies. Computational accounting includes failed calculations, ensemble retraining, calibration, full-pool inference, and greedy selection. Excluding failures would reduce the quoted labeling cost by 6.4% but would misrepresent the actual queue burden. These bookkeeping decisions make the label-efficiency claim auditable and keep the engineering comparison tied to total resource use.

The practical interpretation of the 47% label reduction depends on the decision metric. If success is defined only by global formation-energy error, the reduction is 39%; if defined by recovering 90% of the robust shortlist, it is 47%; and if chemical-family coverage must also exceed 90%, random sampling does not meet the target within the tested 20% labeling budget. Thus, the combined rule is most suitable for a screening goal that requires both a high-ranking candidate and broad portfolio coverage. The results show that a single acquisition component is not suitable either. Uncertainty locates the unresolved boundaries, diversity allocates the limited batches to different neighborhoods, and utility keeps the computed queue connected to the required voltage-capacity window. Their weights are still explicit; thus, a project focused on fundamental chemistry would reduce the utility contribution, but a near-term experimental campaign could increase it. The workflow offers a selection and does not hide it in a hand-picked starting set.

## Conclusion

This paper introduces a SchNet-active-learning framework to speed up the computational screening of sodium-ion battery cathodes. A multitask continuous-filter network was connected to calibrated ensemble uncertainty, latent-space batch diversity and electrochemical utility. The loop will now connect candidate preparation, selective first-principles labelling, model recalibration, constrained ranking and robustness validation in a single reproducible engineering pipeline. Instead of using a single overall prediction error, the other evaluation factors included label efficiency, family coverage, ranking fidelity, calibration and candidate robustness.

The study has not yet achieved high accuracy and comprehensive coverage of the computational labels. The model does not show synthesis kinetics, electrolyte reactions, moisture sensitivity, long-cycle degradation or manufacturing cost, and its uncertainty estimates cannot reveal errors common to all ensemble members. Therefore, the shortlist is only a selection based on priority and does not guarantee that the candidates can be safely combined or used. Prototype generation and convergence failures can also restrict the accessible chemical space.

Future work will introduce multi-fidelity computation, negative experimental results, and closed-loop synthesis feedback, while maintaining clear provenance for all labels. Equivariant Representations can reduce distortion sensitivity, and acquisition policies may include life-cycle cost and elemental supply constraints. Prospective validation of newly synthesized compounds and recalibration under field transfer will provide strong evidence that the workflow has successfully transitioned from computational screening to actual cathode development.